\documentclass[14pt]{article}

\usepackage{INTERSPEECH2019}
\usepackage{hyperref}
\usepackage[super]{nth}

\title{ASSERT: Anti-Spoofing with Squeeze-Excitation and Residual neTworks}
\name{Cheng-I Lai, Nanxin Chen, Jes\'us Villalba, Najim Dehak}

\address{Center for Language and Speech Processing, Johns Hopkins University, Baltimore, MD, USA}
\email{$\{$clai24,bobchennan,jvillal7,ndehak3$\}$@jhu.edu}

\begin{document}

\maketitle
\begin{abstract}
    %(Limit to 200 words)
    We present JHU's system submission to the ASVspoof 2019 Challenge: Anti-Spoofing with Squeeze-Excitation and Residual neTworks (ASSERT).  
    Anti-spoofing has gathered more and more attention since the inauguration of the ASVspoof Challenges, and ASVspoof 2019 dedicates to address attacks from all three major types: text-to-speech, voice conversion, and replay. 
    Built upon previous research work on Deep Neural Network (DNN), ASSERT is a pipeline for DNN-based approach to anti-spoofing.
    ASSERT has four components: feature engineering, DNN models, network optimization and system combination, where the DNN models are variants of squeeze-excitation and residual networks.
    We conducted an ablation study of the effectiveness of each component on the ASVspoof 2019 corpus, and experimental results showed that ASSERT obtained more than $93\%$ and $17\%$ relative improvements over the baseline systems in the two sub-challenges in ASVspooof 2019, ranking ASSERT one of the top performing systems. 
    Code and pretrained models will be made publicly available.
    
\end{abstract}

\section{Introduction}

    % give a brief introduction on anti-spoofing (audio replay + TTS/VC 
    Automatic Speaker Verification (ASV) has become an increasingly attractive option for biometric authentication.
    Past research has shown that ASV systems are subject to malicious attacks: the presentation attacks. 
    Presentation attacks, or spoofing attacks, refer to attempts of bypassing ASV systems by mimicking the voice characteristics of the target speaker. 
    Spoofing attacks have four widely-recognized specifications: impersonation, replay, text-to-speech (TTS) and voice conversion (VC). 
    To defend against these attacks, a standalone anti-spoofing system is developed in parallel to the ASV system~\cite{wu2015spoofing}.  
    Recent efforts on anti-spoofing developments mainly originated from the Biennial ASVspoof Challenges~\cite{evans2013spoofing,wu2015asvspoof,delgado2018asvspoof}.
    
    % introduce ASVspooof 2019. Logical Access (LA) and Physical Access (PA)) 
    Previous ASVspoof Challenges focused on promoting awareness and fostering solutions to spoofing attacks generated from TTS, VC and replay~\cite{evans2013spoofing,wu2015asvspoof,delgado2018asvspoof}. 
    ASVspoof 2019 aims to address all previous attacks and further extended previous editions of ASVspoof in three aspects:  
    \begin{itemize}
      \item Update attacks with TTS and VC with state-of-the-art technologies, especially those based on neural networks.
      \item Create a more controlled setup for replay attacks, covering acoustic and microphone conditions and predefined replay device qualities. 
      \item Adopt an evaluation metric to assess impacts of standalone anti-spoofing systems to a fixed ASV system. 
    \end{itemize} 
    ASVspoof 2019 Challenge is composed of two sub-challenges: Physical Access (PA) and Logical Access (LA).
    LA considers spoofing attacks generated with TTS and VC, and PA refers to spoofing attacks from replay.
    
    %% introduce ASSERT 
    % why do we propose ASSERT 
    % what is ASSERT 
    % what is unique about ASSERT 
    Research work on anti-spoofing can be divided into one of the three categories: Feature Learning~\cite{todisco2016new,sahidullah2015comparison,alam2018boosting,saranyadecision,suthokumar2018modulation,sailor2018auditory,chen2010speaker,leon2012synthetic,wu2012detecting,wu2013synthetic}, Statistical Modeling~\cite{delgado2018asvspoof,adiban2017sut,wu2012study,khoury2014introducing}, and Deep Neural Network (DNN)~\cite{chen2015robust,lavrentyeva2017audio,lai2018attentive,valenti2018end,chen2017resnet,chettri2018study,shim2018replay,qian2016deep}.
    Having witnessed the successes of DNNs in ASVspoof 2017, we decided to explore and extend several DNN-based systems for the ASVspoof 2019 Challenge. 
    Our objective is to identify and design core components of a working pipeline for DNN-based approach to anti-spoofing. 
    These components, feature engineering, DNN models, network optimization, and system fusion, make up our anti-spoofing system, which we term Anti-Spoofing with Squeeze-Excitation and Residual neTworks, or ASSERT.
    The main contribution of this paper is two-fold:
    \begin{enumerate}
      \item \label{item:one} We conducted experiments on the effectiveness of several DNN models in detecting spoofing attacks generated from audio replay, TTS and VC.  
      The DNN models are based on variants of  Squeeze-Excitation Network (SENet)~\cite{hu2018squeeze} and ResNet~\cite{he2016deep}. 
      To our knowledge, we were the first to introduce SENet and ResNet with statistical pooling to address anti-spoofing, and we also extended our previous work in~\cite{lai2018attentive} such that the DNNs are deeper but faster-trained. 
      \item \label{item:two} We presented an ablation study, from feature engineering, network optimization, to fusion schemes for training DNN models for anti-spoofing.  
      We believe these collective strategies are vital for the performance of DNNs. 
      In addition, we compared ASSERT with our implementation of i-vectors baselines~\cite{dehak2011front}. 
      Results on the ASVspoof 2019 corpus demonstrated that ASSERT achieved significant performances over the baseline systems, with more than $93\%$ and $17\%$ relative improvements on PA and LA respectively.  
      Our fusion system was ranked \nth{3} in the PA sub-challenge, and \nth{14} in the LA sub-challenge.
    \end{enumerate}
    
    The outline of the paper is organized as follows. 
    Section~\ref{sec:assert} details ASSERT, from the feature engineering approaches, proposed DNN models, to the optimization and fusion schemes. 
    Section~\ref{sec:exp} compares the results of ASSERT with the baseline systems on the ASVspoof 2019 corpus.  
    We ended the paper with some concluding remarks in Section~\ref{sec:conclusion}.

%%%%%%%%%%%%%%%%%%%%%%%%%%%%%%%%%%%%%%%%%%%%%%%%%%%%%%%%
    
    \begin{figure}[tb]
     \centering
      \includegraphics[width=0.9\columnwidth]{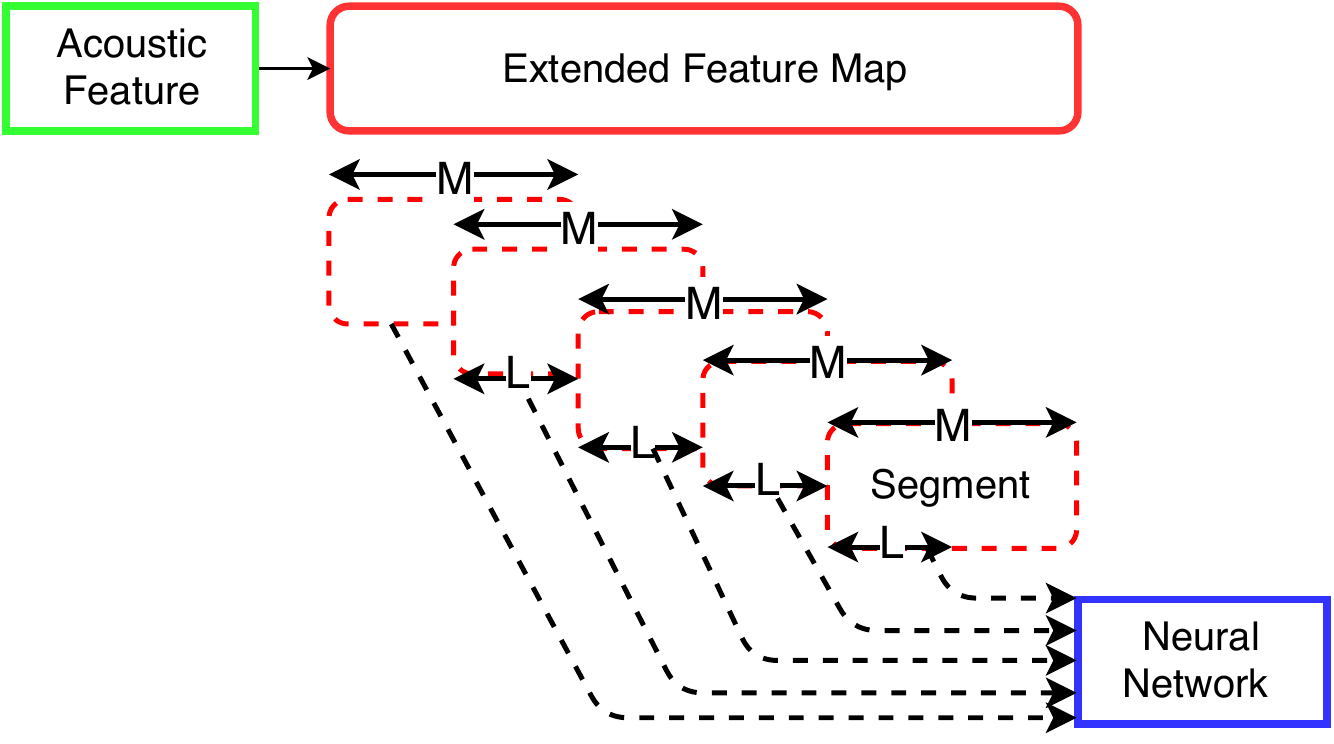}
    \caption{Illustration of Unified Feature Map approach. 
    Low-level acoustics feature are first extracted, and the utterance is repeated to form a unified feature map.
    Then, the feature map is broken down into segments with length $M$ frames and overlap $L$ frames, before inputting into the DNN models.}
    \label{fig:unified_feature_map}
    \end{figure}

\section{ASSERT}
\label{sec:assert}
    This section presents an overview of each component of ASSERT: input feature representations to DNN models, the DNN models and their parameters, along with the network optimization and fusion schemes.  
    The feature preparation is either a unified feature map or the whole utterance. 
    Both approaches are based on some low-level acoustic features.
    The DNN models are variants of squeeze-excitation and residual networks: SENet34, SENet50, Mean-Std ResNet, Dilated ResNet, and Attentive-Filtering Network. 

\subsection{Feature Engineering}
\label{sec:feature_eng}

    \noindent\textbf{Acoustic Features:} We extracted two different acoustic features: constant Q cepstral coefficients (CQCC)~\cite{todisco2016new} and log power magnitude spectra (logspec). 
    Following~\cite{delgado2018asvspoof}, we extracted 30 dimension CQCC feature, including the 0'th order cepstral coefficient and without CMVN. 
    The dimension of logspec is 257.
    For both CQCC and logspec, we \textbf{did not} apply voice activity detection nor any normalization to the acoustic features, as we empirically found doing so yield better results.

    \vspace{1mm}
    \noindent\textbf{Unified Feature Map\footnote{This is not practical for long utterances, but spoofed speech are mostly recorded in short duration (less than 10 seconds).}:} We followed previous work~\cite{lai2018attentive} and created a unified feature map as input to the DNN models. 
    Since the lengths of evaluation utterances were not known beforehand, we first extended all utterances to multiple of $M$ frames. 
    Then, the extended feature map was broken down into segments of length $M$ frames. The segments can have overlap $L$ frames.
    
    For the 2019 ASVspoof Challenge, $M$ is set to 400, and $L$ is set to either 0 or 200.
    Figure~\ref{fig:unified_feature_map} is an illustration of this feature engineering approach. 
    There may be multiple segments per utterance. We simply averaged the DNN outputs over all segments for each utterance.
    
    \vspace{1mm}
    \noindent\textbf{Whole Utterance:} In addition to the Unified Feature Map, we considered another feature engineering approach by training models with the whole utterance (variable length input).
    For each minibatch during training, utterances are zero-padded to match the length of the longest utterance.
    Padding frames are subsequently removed in the pooling layer of the DNNs.

    \begin{figure}[tb]\centering
    \includegraphics[width=.8\columnwidth]{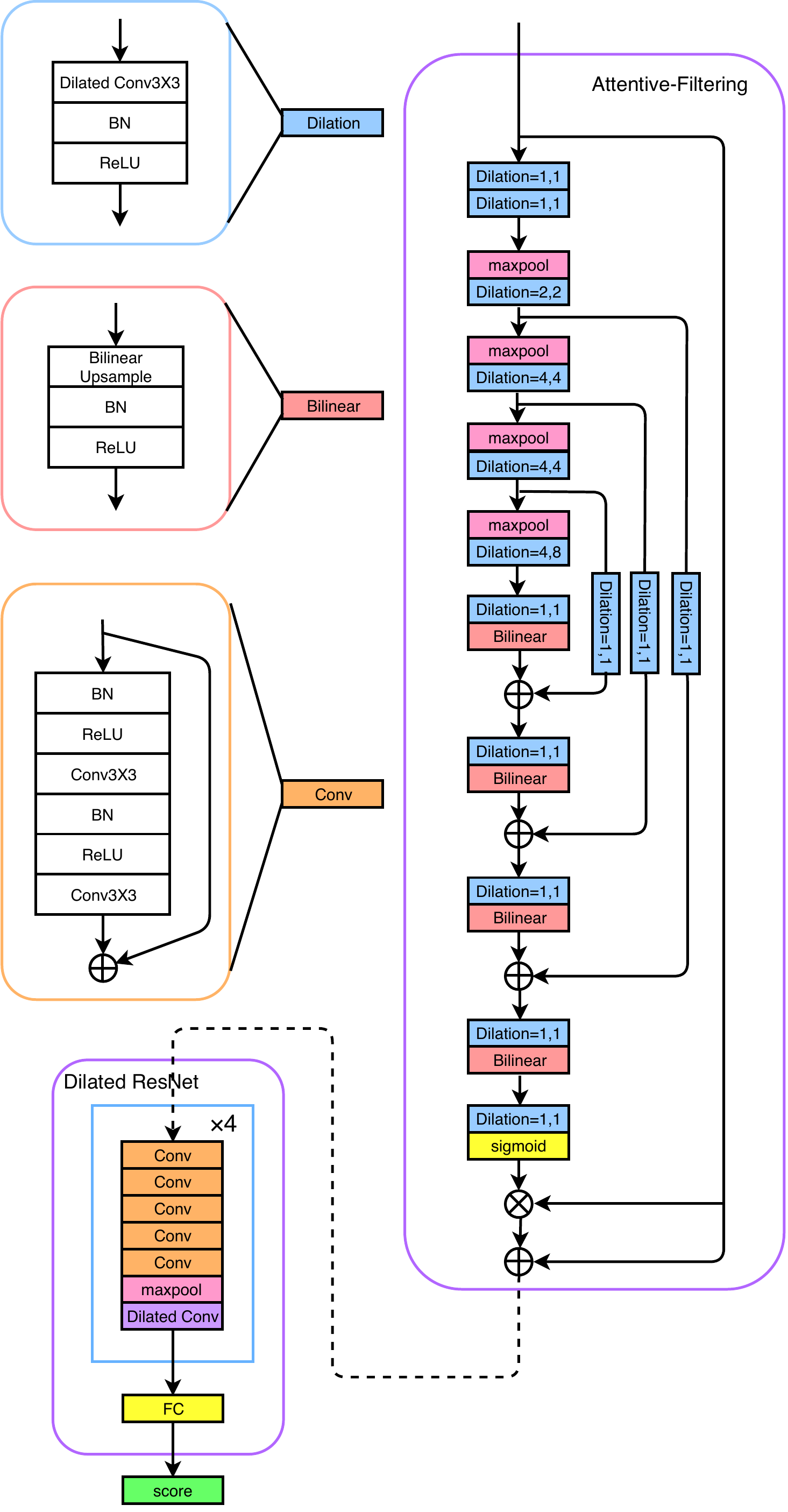}
        \caption{\textbf{(Bottom Left)} Dilated ResNet is consisted of four blocks followed by a fully-connected layer. 
        Each block has five residual units, a max-pooling layer, and a dilated convolution layer.
        \textbf{(Right)} Attentive-Filtering applies an attention-based masking prior to a Dilated ResNet. 
        Input feature goes through four downsampling and four upsampling units. 
        Skip connection is used throughout. 
        Dilation indicates the dilation rate of each convolution layer. 
        Bilinear indicates bilinear upsampling.}
    \label{fig:dilated+attentive_filteirng_resnet_architecture}
    \end{figure}

    \begin{table}[tb]\centering
      \caption{Model parameters of SENet34, SENet50, Mean-Std ResNet, and Dilated ResNet. 
      BN stands for a bottleneck residual block. 
      Basic and Bottleneck residual blocks are described in the original ResNet~\cite{he2016deep}.}
      \label{tbl:model_parameter_summary}
      \resizebox{\columnwidth}{!}{
      \begin{tabular}{@{}clcccc@{}}
        \toprule
        Model & Config. & Block1 & Block2 & Block3 & Block4 \\
        \midrule
        & unit type & Basic & Basic & Basic & Basic \\
        SENet34 & num. of unit & 3 & 4 & 6 & 3 \\ 
        & channels & 16 & 32 & 64 & 128 \\ 
        \midrule
        & unit type & BN & BN & BN & BN \\
        SENet50 & num. of unit & 3 & 4 & 6 & 3 \\ 
        & channels & 16 & 32 & 64 & 128 \\ 
        \midrule
        & unit type & Basic & Basic & Basic & Basic \\ 
        Mean-Std & num. of unit & 3 & 4 & 6 & 3 \\ 
        ResNet & dilation rate & 1 & 1 & 1 & 1 \\
        & channels & 16 & 32 & 64 & 128 \\ 
        \midrule
        & unit type & Basic & Basic & Basic & Basic \\ 
        Dilated & num. of unit & 5 & 5 & 5 & 5 \\ 
        ResNet & dilation rate & 2 & 4 & 4 & 8 \\ 
        & channels & 8 & 16 & 32 & 64 \\ 
        \bottomrule
      \end{tabular}
    }
    \end{table}

\subsection{DNN model}
\label{sec:dnn_model}
    \noindent\textbf{Squeeze-Excitation Network:} Given recent achievements in spoofing countermeasures from different DNN architectures~\cite{lavrentyeva2017audio,lai2018attentive}, we explored an extension of ResNet, Squeeze-Excitation Network (SENet), for ASVspoof 2019.
    SENet has attained impressive image classification results, where a channel-wise transform is appended to existing DNN building blocks, such as the Residual unit~\cite{hu2018squeeze}. 
    We implemented two variants of SENets: SEnet34 with ResNet34 backbone, and SEnet50 with ResNet50 backbone.
    Table~\ref{tbl:model_parameter_summary} contains the model parameters of SEnet34 and SEnet50. 
    SEnet34 and SEnet50 were trained with unified feature maps of logspec while each minibatch contains 64 feature maps. 

    \vspace{1mm}
    \noindent\textbf{Mean-Std ResNet:} Recent work in speaker recognition~\cite{cai2018novel, villalbajhu} has demonstrated that ResNet~\cite{he2016deep} with pooling achieves comparable results as x-vectors~\cite{snyder2018x}. 
    Therefore, we introduced ResNet with pooling for anti-spoofing. 
    Specifically, we employed Mean-Std ResNet, where mean and standard deviation are estimated over timesteps to represent the whole utterance~\cite{snyder2018x} after frame-level features are extracted from a ResNet34.
    Table~\ref{tbl:model_parameter_summary} contains the model parameters of a Mean-Std ResNet. 
    Since the pooling layer accounts for variable length input, we train Mean-Std ResNet with the whole utterance. Both CQCC and logpsec were used, while each minibatch contains 64 and 32 full utterances, respectively.

    \vspace{1mm}
    \noindent\textbf{Dilated ResNet:}
    Following previous work~\cite{lai2018attentive}, we applied Dilated ResNet to ASVspoof 2019.
    Different from Mean-Std ResNet, Dilated ResNet contains a dilated convolution layer in each residual block~\cite{yu2015multi}. We also extended the original dilated residual block to multiple residual units.
    Figure~\ref{fig:dilated+attentive_filteirng_resnet_architecture} is a sketch of the Dilated ResNet, and Table~\ref{tbl:model_parameter_summary} contains its model parameters.
    Contrary to Mean-Std ResNet, Dilated ResNet does not have any pooling layer and thus only accepts fixed-size input.
    We trained Dilated ResNet with the same condition as the SENets.  

    % \begin{figure}[tb]
    %  \centering
    %   \includegraphics[width=0.4\columnwidth]{figures/dilated-resnet-1.pdf}
    % \caption{Dilated ResNet is consisted of four blocks followed by a fully-connected layer. 
    % Each block has five residual units, a max-pooling layer, and a dilated convolution layer.}
    % \label{fig:dilated_resnet_architecture}
    % \end{figure}
    
    % \begin{figure}[tb]\centering
    %   \includegraphics[width=0.7\columnwidth]{figures/attentive-filtering-network-1.pdf}
    % \caption{Attentive-Filtering applies an attention-based masking prior to a Dilated ResNet. 
    % Input feature goes through four downsampling and four upsampling units. 
    % Skip connection is used throughout. 
    % Dilation indicates the dilation rate of each convolution layer. 
    % Bilinear indicates bilinear upsampling.}
    % \label{fig:afn_architecture}
    % \end{figure}

    \vspace{1mm}
    \noindent\textbf{Attentive-Filtering Network:}
    Lastly, we applied Attentive-Filtering Network to ASVspoof 2019, in which an attention-based feature masking is applied prior to a Dilated ResNet~\cite{lai2018attentive}.
    The feature masking is comprised of four downsampling and four upsampling units. 
    The downsampling unit is based on max-pooling and dilated convolution layers, while the upsampling unit is based on convolution and bilinear upsampling layers. 
    The number of channels for all convolution is set to 8, and the final non-linearity is selected as the sigmoid function. 
    Skip connections are included between downsampled and upsampled intermediate representations.
    Furthermore, we also extended the original attentive-filtering proposed in~\cite{lai2018attentive} by replacing the bilinear upsampling method by transpose convolution and self-attention~\cite{vaswani2017attention}. 
    We only reported results with bilinear upsampling in this paper. 
    Figure~\ref{fig:dilated+attentive_filteirng_resnet_architecture} is a visualization of the composition of Attentive-Filtering Network.
    Attentive-Filtering Network was trained under the same condition as the SENets. 
    
\subsection{Network Optimization}
\label{sec:optimization}
    We present the optimization schemes used for training all the DNN models described above.
    
    \vspace{1mm}
    \noindent\textbf{Training Objective:}
    The objective of anti-spoofing is to classify whether an utterance is bonafide or spoofed. 
    A straight forward training objective is, therefore, binary classification~\cite{lavrentyeva2017audio,lai2018attentive}.     
    ~\cite{shim2018replay} showed that it could be beneficial to optimize the networks by classifying noise classes in audio replay, so we further trained our models with multi-class classification.
    We designed the multi-class labels as the conditions spoofed utterances are generated. 
    For LA, this is simply the system ID (bonafide, SS\_1, SS\_2, SS\_4, US\_1, VC\_1, VC\_4) e.g. SS\_1 corresponds to ``system using neural-network acoustic models and WaveNet vocoder."
    For PA, spoofed utterances are recorded under different environment IDs and attack IDs.
    In this work, we adopted attack ID (bonafide, AA, AB, AC, BA, BB, BC, CA, CB, CC) as the multi-class label. 
    Table~\ref{tbl:multi_class_label} gives a comparison between the multi-class classification labels between condition LA and PA. 
    For computing EER during inference stage, we took the log-probability of the bonafide class as the score for utterance. 
    
    \begin{table}[tb]\centering
    \caption{Multi-class labels for LA and PA.}
    \label{tbl:multi_class_label}
    \begin{tabular}{l|cc}
        \toprule
        data condition & LA & PA \\ \hline
        num. of classes & 7 & 10 \\ \hline
        label type & system IDs & attack IDs \\ \hline
        \begin{tabular}[l]{@{}l@{}}label \\ descriptions\end{tabular} & \begin{tabular}[c]{@{}c@{}c@{}} bonafide, SS\_1 \\ SS\_2,  SS\_4 \\ US\_1, VC\_1, VC\_4\end{tabular} & \begin{tabular}[c]{@{}c@{}c@{}} bonafide, AA, AB \\ AC, BA, BB \\ BC, CA, CB, CC\end{tabular} \\
        \bottomrule
    \end{tabular}
    \end{table}

    \vspace{1mm}
    \noindent\textbf{Optimizer:}
    We followed the optimization strategy described in~\cite{vaswani2017attention}.
    The DNN models are optimized by Adam with $\beta_{1} = 0.9$,  $\beta_{2} = 0.98$, and weight decay $10^{-9}$.
    The learning rate scheduler increases the learning rate linearly for the first $1000$ warm-up steps and then decreases it proportionally to the inverse square root of the step number~\cite{vaswani2017attention}.
    Finally, after every training epoch, we selected the best model by two means: \textit{dev} EER or \textit{dev} classification accuracy of the class labels. 

\subsection{Fusion and Calibration}
\label{sec:fusion}
    We followed the greedy fusion scheme described in~\cite{villalbajhu} to select the best system combination for our primary submission for ASVspoof 2019.
    Fusion and calibration were performed with logistic regression with the Bosaris toolkit~\cite{brummer2013bosaris}.
    Considering the priors defined in the evaluation plan, we set the effective prior to $0.672$ for PA and the effective prior to $0.707$ for LA.

%%%%%%%%%%%%%%%%%%%%%%%%%%%%%%%%%%%%%%%%%%%%%%%%%%%%%%%%

   \begin{table*}[t]
      \caption{Ablation study of single system results on ASVspoof 2019. 
      Due to space constraint, for each DNN model, we merely included \textbf{top two} performing systems in the table. 
      Under Training Objective column, MCE stands for multi-class cross entropy, BCE stands for binary cross entropy, and acc/EER  stands for the model selection criterion after each training epoch.}
      \label{tab:single_system_results}
      \centering
      \begin{tabular}{@{}lcccccccc@{}}
        \toprule
        Model & Acoustic & Feature & Training & Model & \multicolumn{2}{c}{PA development} & \multicolumn{2}{c}{LA development} \\
        % \cmidrule(rl){6-7}
        % \cmidrule(rl){8-9}
        & Feature & Engineering & Objective & Params. & t-DCF$_{norm}^{min}$ & EER ($\%$) & t-DCF$_{norm}^{min}$ & EER ($\%$) \\
        \midrule
        CQCC-GMM & CQCC + $\Delta$ + $\Delta\Delta$ & N/A & EM & 138k & 0.195 & 9.87 & 0.012 & 0.43 \\
        LFCC-GMM & LFCC + $\Delta$ + $\Delta\Delta$ & N/A & EM & 92k & 0.255 & 11.96 & 0.066 & 2.71 \\
        100-i-vectors & CQCC + $\Delta$ + $\Delta\Delta$ & N/A & EM & 593k & 0.306 & 12.37 & 0.155 & 5.18 \\
        200-i-vectors & CQCC + $\Delta$ + $\Delta\Delta$ & N/A & EM & 2339k & 0.322 & 12.52 & 0.121 & 4.12 \\
        \midrule
        SENet34 & logspec & unifed, L=200 & BCE + acc. & 1344k & \bf 0.015 & \bf 0.575 & \bf 0 & \bf 0 \\
        & logspec & unifed, L=200 & BCE + EER & 1344k & 0.017 & 0.686 & 0 & 0 \\
        \midrule
        SENet50 & logspec & unifed, L=200 & MCE + EER & 1095k & 0.021 & 0.799 & 0 & 0 \\
        & logspec & unifed, L=200 & BCE + EER & 1093k & 0.017 & 0.631 & 0 & 0 \\       
        \midrule
        Mean-Std & logspec & whole & BCE + acc. & 1389k & 0.022 & 0.832 & 0 & 0 \\
        ResNet & CQCC & whole & MCE + acc. & 1390k & 0.041 & 1.429 & 0.001 & 0.040 \\    
        \midrule
        Dilated & logspec & unifed, L=200 & MCE + EER & 593k & 0.029 & 1.072 & 0 & 0 \\
        ResNet & logspec & unifed, L=200 & BCE + EER & 592k & 0.024 & 0.780 & 0 & 0 \\ 
        \midrule
        Attentive- & logspec & unifed, L=200 & MCE + EER & 600k & 0.027 & 1.057 & 0 & 0 \\
        Filteirng Net & logspec & unifed, L=200 & BCE + acc. & 599k & 0.021 & 0.740 & 0 & 0 \\ 
        
        \bottomrule
      \end{tabular}
    \end{table*}

\section{Experiments}
\label{sec:exp}

\subsection{Baseline System}
\label{sec:exp_baseline}
    \noindent\textbf{LFCC-GMM and CQCC-GMM:} The official baseline systems are based on a 20 dimension linear frequency cepstral coefficients (LFCC)~\cite{sahidullah2015comparison} and a 30 dimension constant Q cepstral coefficients (CQCC)~\cite{todisco2016new}, both of which included static, delta and double delta coefficients. 
    The backend is a 2-class GMM with 512 components.
    
    \vspace{1mm}
    \noindent\textbf{i-vectors:} We implemented i-vectors~\cite{dehak2011front} as previous work has demonstrated that i-vectors yield better results than GMM in anti-spoofing~\cite{delgado2018asvspoof,adiban2017sut}.
    We experimented with a 100-dimension i-vectors with 64-component UBM and a 200-dimension i-vectors with 128-component UBM. 
    The i-vector extractors were trained on 30-dimension CQCC features, and the backend for our i-vectors is a Gaussian linear generative model~\cite{martinez2011language}.
    
\subsection{Experimental Setup}
\label{sec:experimental_setup}
    \noindent\textbf{Dataset:} All experiments in this work were conducted under the ASVspoof 2019 Challenge.
    The ASVspoof 2019 corpus is divided into two subsets: PA and LA. 
    PA contains 48600 spoof and 5400 bonafide utterances in the training partition, 24300 spoof and 5400 bonafide utterances in the development partition. 
    The spoofed utterances were recorded under are 27 different acoustic and 9 replay configurations. 
    LA contains 22800 spoof and 2580 bonafide utterances in the training partition, 22296 spoof and 2548 bonafide utterances in the development partition. 
    The spoofed utterances were generated according to 2 VC and 4 TTS algorithms.
    
    \vspace{1mm}
    \noindent\textbf{Evaluation Metric:} We evaluated the effectiveness of spoofing countermeasures with the \textit{minimum} normalized tandem detection cost function (t-DCF)~\cite{kinnunen2018t} and EER.
    t-DCF takes into account of the detection error rates of a fixed automatic speaker verification system, provided by the organizers in the case of ASVspoof 2019 Challenge. 
    
    \vspace{1mm}
    \noindent\textbf{Implementation:} We used training partition to train our DNN models. Development partition was used for model selection during validation and system combination. We \textbf{did not} use any external data or data augmentation technique for development. CQCC-GMM and LFCC-GMM were adopted directly from the MATLAB script\footnote{\href{http://www.asvspoof.org}http://www.asvspoof.org}. 
    Acoustic features and i-vectors were extracted with Kaldi~\cite{povey2011kaldi}. 
    DNNs were implemented in PyTorch. 

\subsection{Experimental Results of Single Systems}
\label{sec:exp_single_results}

    Table~\ref{tab:single_system_results} compares ASSERT with the baseline systems in spoofing countermeasure on the \textit{dev} partition. 
    The first observation is that the i-vectors baseline performed worse than GMMs surprisingly, which is contrary to prior work on the ASVspoof 2017 corpus~\cite{delgado2018asvspoof,adiban2017sut}. 
    ASSERT attains substantial improvements from the baseline GMM and i-vectors systems on both PA and LA.
    In general, for training the proposed DNN models, logspec outperforms CQCC, and unified feature map with overlap outperforms without overlap and whole utterance. 
    On the other hand, there are mixed results on using multi-task or binary training objective, and on model selection with \textit{dev} EER or \textit{dev} classification accuracy.
    We empirically found that the best single system is based on SENet34 trained with unified feature map with overlap of logspec and binary cross-entropy loss with \textit{dev} accuracy model selection.
    The system obtains 92\% and 94\% relative improvements over CQCC-GMM on \textit{dev} t-DCF and EER for PA, and 100\% relative improvements for LA. 

\subsection{Evaluation Results}
\label{sec:exp_fusion_results}
    
    Table~\ref{tab:submission_results} is the summary of our primary and single system submission to the ASVspoof 2019 Challenge. 
    The single system is based on SENet34 (logspec), and the primary system is a system combination of five single systems based on SENet34 (logspec), Mean-Std ResNet (CQCC, logspec), SENet50 (logspec) and Dilated ResNet (logspec).
    Systems are trained separately for PA and LA.
    We can observe that ASSERT generalizes well across \textit{dev} and \textit{eval} for PA, nevertheless, it overfits on \textit{dev} for LA. 
    Our primary system further gains 93\% and 95\% relative improvements over CQCC-GMM on \textit{eval} t-DCF and EER for PA, and 27\% and 17\% relative improvements over LFCC-GMM on \textit{eval} t-DCF and EER for LA.
    
    \begin{table}[t]
      \caption{Primary, single and baseline systems for ASVspoof 2019.
      Single system is based on SENet34; primary system is a fusion of five systems based on: SENet34, Mean-Std ResNet (CQCC, logspec), SENet50 and Dilated ResNet. 
      }
      \label{tab:submission_results}
      \centering
      \begin{tabular}{@{}lcccc@{}}
        \toprule
        System & \multicolumn{2}{c}{Development} & \multicolumn{2}{c}{Evaluation} \\
        \cmidrule(rl){2-3}
        \cmidrule(rl){4-5}
        & t-DCF$_{norm}^{min}$ & EER & t-DCF$_{norm}^{min}$ & EER \\
        \midrule
        PA-single & 0.015 & 0.575 & 0.036 & 1.29 \\
        PA-primary & \bf 0.003 & \bf 0.129 & \bf 0.016 & \bf 0.59 \\
        PA-baseline & 0.195 & 9.87 & 0.245 & 11.04 \\
        \midrule
        LA-single & 0 & 0 & 0.216 & 11.75 \\
        LA-primary & \bf 0 & \bf 0 & \bf 0.155 & \bf 6.70 \\
        LA-baseline & 0.066 & 2.71 & 0.212 & 8.09 \\
        \bottomrule
      \end{tabular}%}
    \end{table}

%%%%%%%%%%%%%%%%%%%%%%%%%%%%%%%%%%%%%%%%%%%%%%%%%%%%%%%%%%%%%%%%%%%%%%%%%%%%%%%%%%%%%%%%%%%%%%%%%%%%%%%%%%%%%%%%%%%%

\section{Conclusions}
\label{sec:conclusion}
    We introduced ASSERT -- several variants of squeeze-excitation and residual networks, optimization and fusion schemes, along with feature engineering approaches -- for anti-spoofing. 
    Our fusion system attained considerable improvement over baseline systems on the ASVspoof 2019 corpus. 
    We believe this paper serves as a preliminary work on a more comprehensive study on DNN based countermeasures for speech spoofing attacks, while meta-data analysis and model refinements on LA should be further investigated. 
    
    \vspace{1mm}
    \noindent\textbf{Acknowledgments}
    \label{sec:acknowledgment}
    The authors thank ASVspoof 2019 committee for designing the corpus and organizing the challenge. 

%%%%%%%%%%%%%%%%%%%%%%%%%%%%%%%%%%%%%%%%%%%%%%%%%%%%%%%%

% \vfill\pagebreak

\bibliographystyle{IEEEtran}
\bibliography{main}
\end{document}